\documentclass[runningheads]{llncs}
\usepackage{graphicx}
\usepackage[noend]{algorithm2e}
\begin{document}
\title{Model-agnostic multi-objective approach for the evolutionary discovery of mathematical models}

\titlerunning{Multi-objective discovery of mathematical models}
%
\author{Alexander Hvatov\inst{1}\and
Mikhail Maslyaev\inst{1}\and
Iana S. Polonskaya\inst{1}\and
Mikhail Sarafanov\inst{1}\and
Mark Merezhnikov\inst{1}\and
Nikolay O. Nikitin\inst{1}}
\authorrunning{A. Hvatov et al.}
%
\institute{NSS (Nature Systems Simulation) Lab, ITMO University, Saint-Petersburg, Russia\\
\email{\{alex\_hvatov,mikemaslyaev,ispolonskaia, \\mik\_sar,mark.merezhnikov,nnikitin\}@itmo.ru}}
\maketitle              
\begin{abstract}
  In modern data science, it is often not enough to obtain only a data-driven model with a good prediction quality. On the contrary, it is more interesting to understand the properties of the model, which parts could be replaced to obtain better results. Such questions are unified under machine learning interpretability questions, which could be considered one of the area's raising topics. In the paper, we use multi-objective evolutionary optimization for composite data-driven model learning to obtain the algorithm's desired properties. It means that whereas one of the apparent objectives is precision, the other could be chosen as the complexity of the model, robustness, and many others. The method application is shown on examples of multi-objective learning of composite models, differential equations, and closed-form algebraic expressions are unified and form approach for model-agnostic learning of the interpretable models.


\end{abstract}
\label{sec:intro}
\section{Introduction}

The increasing precision of the machine learning models indicates that the best precision model is either overfitted or very complex. Thus, it is used as a black box without understanding the principle of the model's work. This fact means that we could not say if the model can be applied to another sample without a direct experiment. Related questions such as applicability to the given class of the problems, sensitivity, and the model parameters' and hyperparameters' meaning arise the interpretability problem \cite{lipton2018mythos}.

In machine learning, two approaches to obtain the model that describes the data are existing. The first is to fit the learning algorithm hyperparameters and the parameters of a given model to obtain the minimum possible error. The second one is to obtain the model structure (as an example, it is done in neural architecture search \cite{elsken2019neural}) or the sequence of models that describe the data with the minimal possible error \cite{olson2016tpot}. We refer to obtaining a set of models or a composite model as a composite model discovery.

After the model is obtained, both approaches may require additional model interpretation since the obtained model is still a black box. For model interpretation, many different approaches exist. One group of approaches is the model sensitivity analysis \cite{saltelli2010variance}. Sensitivity analysis allows mapping the input variability to the output variability. Algorithms from the sensitivity analysis group usually require multiple model runs. As a result, the model behavior is explained relatively, meaning how the output changes with respect to the input changing.

The second group is the explaining surrogate models that usually have less precision than the parent model but are less complex. For example, linear regression models are used to explain deep neural network models \cite{tsakiri2018artificial}. Additionally, for convolutional neural networks that are often applied to the image classification/regression task, visual interpretation may be used \cite{konforti2020inference}. However, this approach cannot be generalized, and thus we do not put it into the classification of explanation methods.

All approaches described above require another model to explain the results. Multi-objective optimization may be used for obtaining a model with desired properties that are defined by the objectives. The Pareto frontier obtained during the optimization can explain how the objectives affect the resulting model's form. Therefore, both the ``fitted'' model and the model's initial interpretation are achieved.

We propose a model-agnostic data-driven modeling method that can be used for an arbitrary composite model with directed acyclic graph (DAG) representation. We assume that the composite model graph's vertices (or nodes) are the models and the vertices define data flow to the final output node. Genetic programming is the most general approach for the DAG generation using evolutionary operators. However, it cannot be applied to the composite model discovery directly since the models tend to grow unexpectedly during the optimization process. 

The classical solution is to restrict the model length \cite{grosan2004evolving}. Nevertheless, the model length restriction may not give the best result. Overall, an excessive amount of models in the graph drastically increases the fitting time of a given composite model. Moreover, genetic programming requires that the resulting model is computed recursively, which is not always possible in the composite model case.

We refine the usual for the genetic programming cross-over and mutation operators to overcome the extensive growth and the model restrictions. Additionally, a regularization operator is used to retain the compactness of the resulting model graph. The evolutionary operators, in general, are defined independently on a model type. However, objectives must be computed separately for every type of model. The advantage of the approach is that it can obtain a data-driven model of a given class. Moreover, the model class change does not require significant changes in the algorithm.

We introduce several objectives to obtain additional control over the model discovery. The multi-objective formulation allows giving an understanding of how the objectives affect the resulting model. Also, the Pareto frontier provides a set of models that an expert could assess and refine, which significantly reduces time to obtain an expert solution when it is done ``from scratch''. Moreover, the multi-objective optimization leads to the overall quality increasing since the population's diversity naturally increases.

Whereas the multi-objective formulation \cite{vu2019toward} and genetic programming \cite{lu2016using} are used for the various types of model discovery separately, we combine both approaches and use them to obtain a composite model for the comprehensive class of atomic models. In contrast to the composite model, the atomic model has single-vector input, single-vector output, and a single set of hyperparameters. As an atomic model, we may consider a single model or a composite model that undergoes the ``atomization'' procedure.  Additionally, we consider the Pareto frontier as the additional tool for the resulting model interpretability.

The paper is structured as follows: Sec.~\ref{sec:problem_statement} describes the evolutionary operators and multi-objective optimization algorithms used throughout the paper, Sec.~\ref{sec:experimental_results} describes discovery done on the same real-world application dataset for particular model classes and different objectives. In particular, Sec.~\ref{sec:FEDOT} describes the composite model discovery for a class of the machine learning models with two different sets of objective functions; Sec.~\ref{sec:EALG} describes the closed-form algebraic expression discovery; Sec.~\ref{sec:EPDE} describes the differential equation discovery. Sec.~\ref{sec:conclusion} outlines the paper.

\label{sec:problem_statement}
\section{Problem statement for model-agnostic approach}

The developed model agnostic approach could be applied to an arbitrary composite model represented as a directed acyclic graph. We assume that the graph's vertices (or nodes) are the atomic models with parameters fitted to the given data. It is also assumed that every model has input data and output data. The edges (or connections) define which models participate in generating the given model's input data.

\begin{figure}[ht!]
\centering
\includegraphics[width=0.9\linewidth]{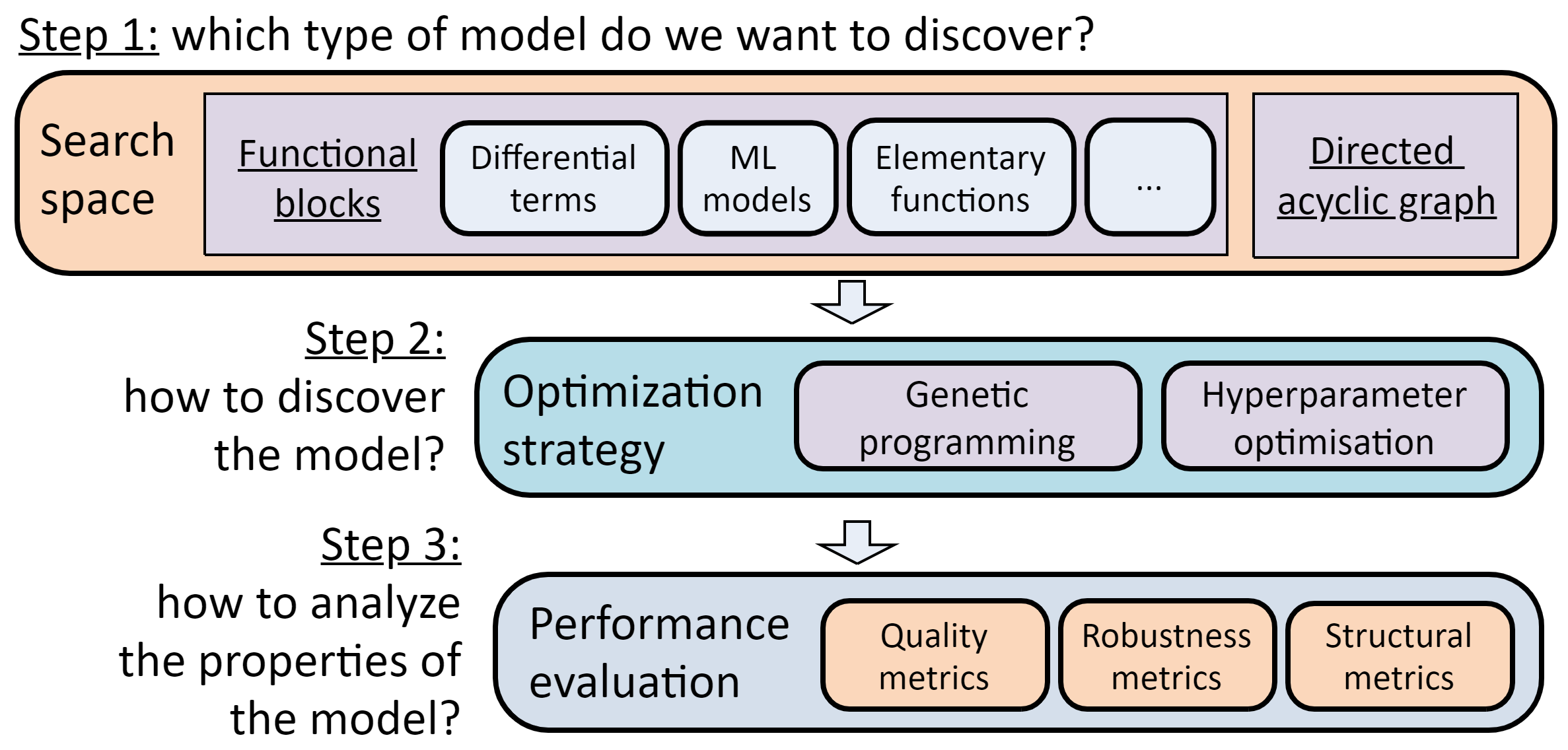}
\caption{Illustration of the modules of the composite models discovery system and the particular possible choices for the realization strategy.}
\label{fig:approach_questions}
\end{figure}

Before the direct approach description, we outline the scheme Fig.~\ref{fig:approach_questions} that illustrates the step-by-step questions that should be answered to obtain the composite model discovery algorithm.

For our approach, we aim to make Step 1 as flexible as possible. Approach application to different classes of functional blocks is shown in Sec.~\ref{sec:experimental_results}. The realization of Step 2 is described below. Moreover, different classes of models require different qualitative measures (Step 3), which is also shown in Sec.~\ref{sec:experimental_results}.

\begin{figure}[ht!]
\centering
\includegraphics[width=\linewidth]{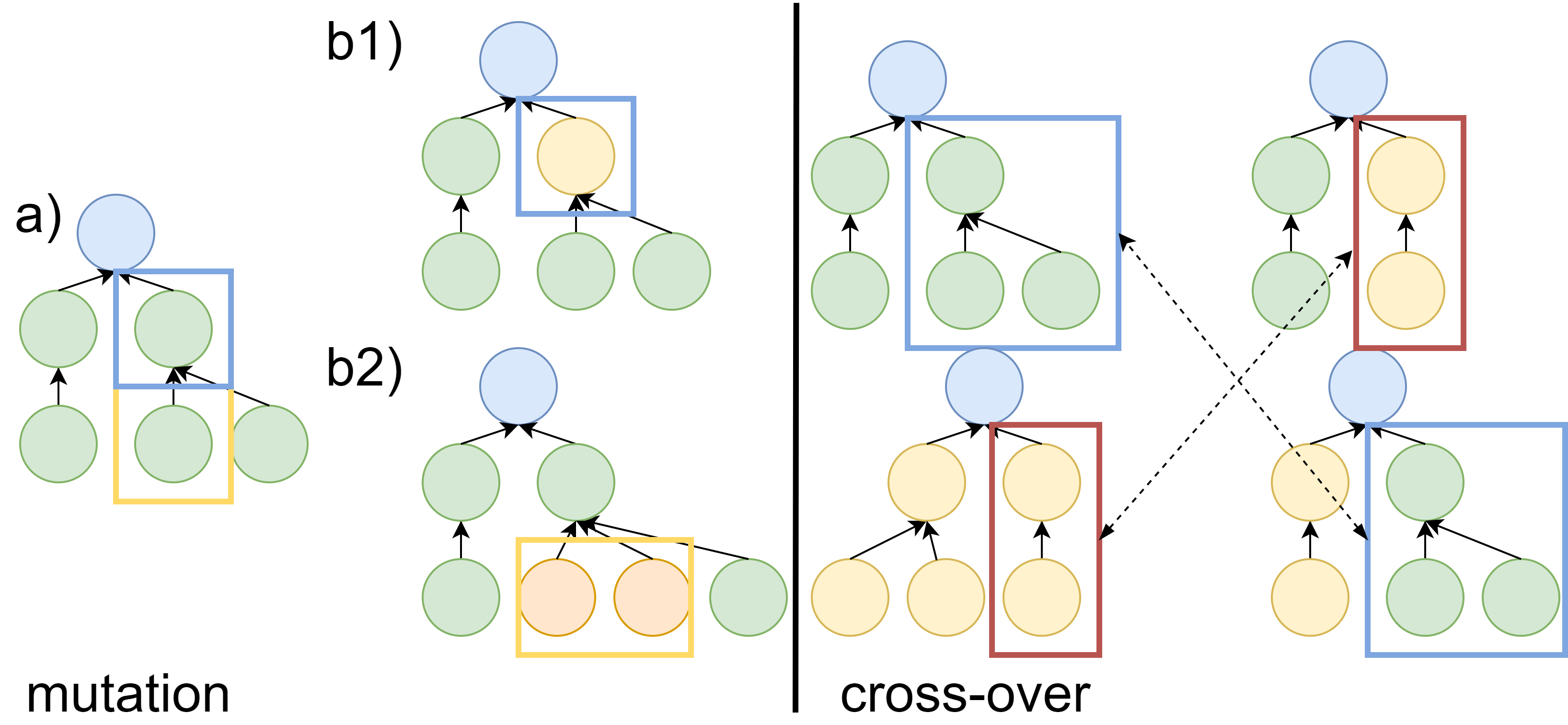}
\caption{ (left) The generalized composite model individual: a) individual with models as nodes (green) and the dedicated output node (blue) and nodes with blue and yellow frames that are subjected to the two types of the mutations b1) and b2); b1) mutation with change one atomic model with another atomic model (yellow); b2) mutation with one atomic model replaced with the composite model (orange) 
(right) The scheme of the cross-over operator green and yellow nodes are the different individuals. The frames are subtrees that are subjected to the cross-over (left), two models after the cross-over operator is applied (right) }
\label{fig:common_mutation_xover_scheme}
\end{figure}

The cross-over and mutation schemes used in the described approach do not differ in general from typical symbolic optimization schemes. However, in contrast to the usual genetic programming workflow, we have to add regularization operators to restrict the model growth. Moreover, the regularization operator allows to control the model discovery and obtain the models with specific properties.  In this section, we describe the generalized concepts. However, we note that every model type has its realization details for the specific class of the atomic models. Below we describe the general scheme of the three evolutionary operators used in the composite model discovery: cross-over, mutation, and regularization shown in Fig.~\ref{fig:common_mutation_xover_scheme}.

The mutation operator has two variations: node replacement with another node and node replacement with a sub-tree. Scheme for the both type of the mutations are shown in Fig.~\ref{fig:common_mutation_xover_scheme}(left). The two types of mutation can be applied simultaneously. The probabilities of the appearance of a given type of mutation are the algorithm's hyperparameters. We note that for convenience, some nodes and model types may be ``immutable''. This trick is used, for example, in Sec.\ref{sec:EPDE} to preserve the differential operator form and thus reduce the optimization space (and consequently the optimization time) without losing generality.

In general, the cross-over operator could be represented as the subgraphs exchange between two models as shown in Fig.~\ref{fig:common_mutation_xover_scheme} (right). In the most general case, in the genetic programming case, subgraphs are chosen arbitrarily. However, since not all models have the same input and output, the subgraphs are chosen such that the inputs and outputs of the offspring models are valid for all atomic models.

\begin{figure}[ht!]
\centering
\includegraphics[width=0.5\linewidth]{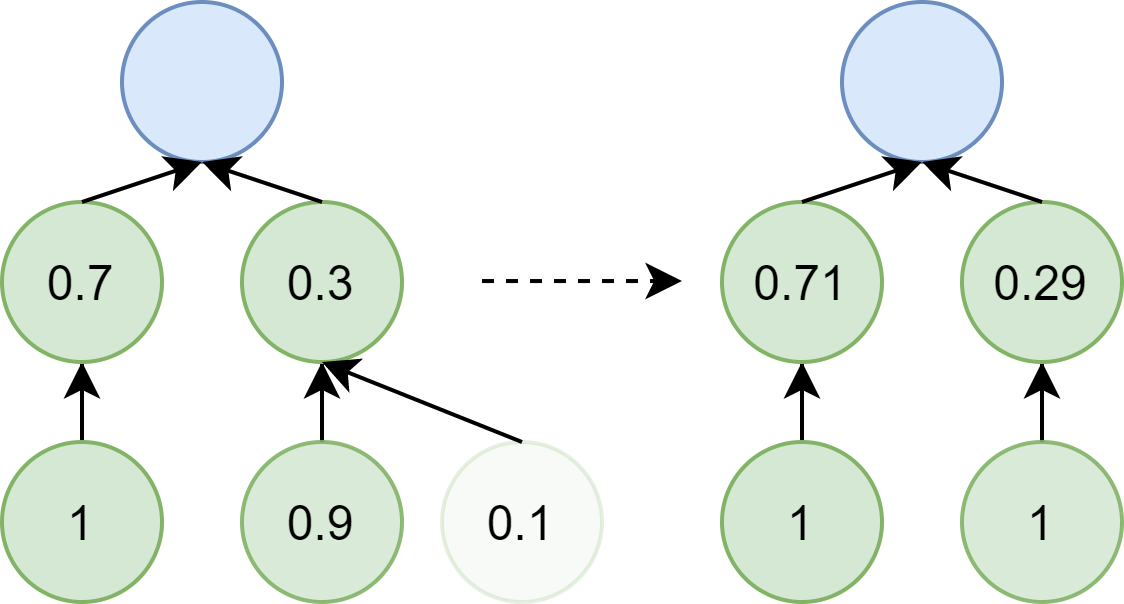}
\caption{The scheme of the regularization operator. The numbers are the dispersion ratio that is described by the child nodes as the given depth level. In the example model with dispersion ratio $0.1$ is removed from the left model to obtain simpler model to the right.}
\label{fig:common_regularization_scheme}
\end{figure}

In order to restrict the excessive growth of the model, we introduce an additional regularization operator shown in Fig.~\ref{fig:common_regularization_scheme}. The amount of described dispersion (as an example, using the $R2$ metric) is assessed for each graph's depth level. The models below the threshold are removed from the tree iteratively with metric recalculation after each removal. Also, the applied implementation of the regularization operator can be task-specific (e.g., custom regularization for composite machine learning models \cite{nikitin2020structural} and the LASSO regression for the partial differential equations \cite{MASLYAEV2021101345}).

Unlike the genetic operators defined in general, the objectives are defined for the given class of the atomic models and the given problem. The class of the models defines the way of an objective function computation. For example, we consider the objective referred to as ``quality'', i.e., the ability of the given composite model to predict the data in the test part of the dataset. Machine learning models may be additionally ``fitted'' to the data with a given composite model structure. Fitting may be used for all models simultaneously or consequently for every model. Also, the parameters of the machine learning models may not be additionally optimized, which increases the optimization time.  The differential equations and the algebraic expressions realization atomic models are more straightforward than the machine learning models. The only way to change the quality objective is the mutation, cross-over, and regularization operators. We note that there also may be a ``fitting'' procedure for the differential equations. However, we do not introduce variable parameters for the differential terms in the current realization.

In the initial stage of the evolution, according to the standard workflow of the MOEA/DD \cite{li2014evolutionary} algorithm, we have to evaluate the best possible value for each of the objective functions. The selection of parents for the cross-over is held for each objective function space region. With a specified probability of maintaining the parents' selection, we can select an individual outside the processed subregion to partake in the recombination. In other cases, if there are candidate solutions in the region associated with the weights vector, we make a selection among them. The final element of MOEA/DD is population update after creating new solutions, which is held without significant modifications. The resulting algorithm is shown in Alg.~\ref{alg:approach}.

\begin{algorithm}[h!]
\label{alg:approach}
 \KwData{Class of atomic models $T = \{T_1, T_2, \; ...\; T_n\}$; (optional) define subclass of immutable models; objective functions}
 \KwResult{Pareto frontier}
 Create a set of weight vectors $\mathbf{w} = (w^1, ..., w^{n\_weights}), \; w^i=(w^i_1, \; ..., \; w^i_{n\_eq + 1}) $\;
 \For{weight\_vector in weights}{
 Select K nearest weight vectors to the weight\_vector\;
 }
 Randomly generate a set of candidate models \& divide them into non-dominated levels\;
 Divide the initial population into groups by subregion, to which they belong\;
 \For{epoch = 1 to epoch\_number}{
  \For{weight\_vector in weights}{
   Parent selection\;
   Apply recombination to parents pool and mutation to individuals inside the region of weights (Fig.~\ref{fig:common_mutation_xover_scheme})\;
   \For{offspring in new\_solutions}{
    Apply regularization operator (Fig.~\ref{fig:common_regularization_scheme})\;
    Get values of objective functions for offspring\;
   }
   Update population\;
  }
 }
 \caption{The pseudo-code of model-agnostic Pareto frontier construction}
\end{algorithm}

To sum up, the proposed approach combines classical genetic programming operators with the regularization operator and the immutability property of selected nodes. The refined MOEA/DD and refined genetic programming operators obtain composite models for different atomic model classes.

\section{Examples}
\label{sec:experimental_results}

In this section, several applications of the described approach are shown. We use a common dataset for all experiments that are described in the Sec.~\ref{sec:dataset_description}. 

While the main idea is to show the advantages of the multi-objective approach, the particular experiments show different aspects of the approach realization for different models' classes. Namely, we want to show how different choices of the objectives reflect the expert modeling. 

For the machine learning models in Sec.~\ref{sec:FEDOT}, we try to mimic the expert's approach to the model choice that allows one to transfer models to a set of related problems and use a robust regularization operator.

Almost the same idea is pursued in mathematical models. Algebraic expression models in Sec.~\ref{sec:EALG} are discovered with the model complexity objective. More complex models tend to reproduce particular dataset variability and thus may not be generalized. To reduce the optimization space, we introduce immutable nodes to restrict the model form without generality loss. The regularization operator is also changed to assess the dispersion relation between the model residuals and data, which has a better agreement with the model class chosen.

While the main algebraic expressions flow is valid for partial differential equations discovery in Sec.~\ref{sec:EPDE}, they have specialties, such as the impossibility to solve intermediate models ``on-fly''. Therefore LASSO regularization operator is used.

\subsection{Experimental setup}
\label{sec:dataset_description}

The validation of the proposed approach was conducted for the same dataset for all types of models: composite machine learning models, models based on closed-form algebraic expression, and models in differential equations.

\begin{figure}[ht!]
\centering
\includegraphics[width=0.8\linewidth]{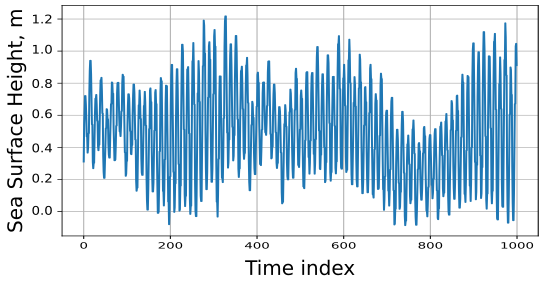}
\caption{The multi-scale time series of sea surface height used as a dataset for all experiments.}
\label{fig_dataset}
\end{figure}

The multi-scale environmental process was selected as a benchmark. As the dataset for the examples, we take the time series of sea surface height were obtained from numerical simulation using the high-resolution setup of the NEMO model for the Arctic ocean \cite{hvatov2019adaptation}. The simulation covers one year with the hourly time resolution. The visualization of the experimental data is presented in Fig.~\ref{fig_dataset}.

It is seen from Fig.~\ref{fig_dataset} that the dataset has several variability scales. The composite models, due to their nature, can reproduce multiple scales of variability. In the paper, the comparison between single and composite model performance is taken out of the scope. We show only that with a single approach, one may obtain composite models of different classes.

\subsection{Composite machine learning models}
\label{sec:FEDOT}

The machine learning pipelines' discovery methods are usually referred to as automated machine learning (AutoML). For the machine learning model design, the promising task is to control the properties of obtained model. Quality and robustness could be considered as an example of the model's properties. The proposed model-agnostic approach can be used to discover the robust composite machine learning models with the structure described as a directed acyclic graph (as described in \cite{nikitin2020structural}). In this case, the building blocks are regression-based machine learning models, algorithms for feature selection, and feature transformation. The specialized lagged transformation is applied to the input data to adapt the regression models for the time series forecasting. This transformation is also represented as a building block for the composite graph \cite{kalyuzhnaya2020automatic}


The quality of the composite machine learning models can be analyzed in different ways. The simplest way is to estimate the quality of the prediction on the test sample directly. However, the uncertainty in both models and datasets makes it necessary to apply the robust quality evaluation approaches for the effective analysis of modeling quality \cite{vychuzhanin2019robust}. 

The stochastic ensemble-based approach can be applied to obtain the set of predictions $Y_{ens}$ for different modeling scenarios using stochastic perturbation of the input data for the model. In this case, the robust objective function $\widetilde{f_i}\left(Y_{ens}\right)$ can be evaluated as follows:

\begin{equation}
\label{eq:ens_metrics}
\begin{array}{cc}
\mu_{ens} = \frac{1}{k}\sum_{j=1}^{k}{(Y_{ens}^j)} + 1, \\
\widetilde{f_i}\left(Y_{ens}\right)=\mu_{ens}\sqrt{\frac{1}{k-1}\sum_{i=1}^{k}\left(f_i\left(Y_{ens}^i\right)-\mu_{ens}\right)^2} + 1
\end{array}
\end{equation}

In Eq.~\ref{eq:ens_metrics} $k$ is the number of models in the ensemble, $f$ - function for modelling error, $Y_{ens}$ - ensemble of the modelling results for specific configuration of the composite model.

The robust and non-robust error measures of the model cannot be minimized together. In this case, the multi-objective method proposed in the paper can be used to build the predictive model. We implement the described approach as a part of the FEDOT framework\footnote{\url{https://github.com/nccr-itmo/FEDOT}} that allow building various ML-based composite models. The previous implementation of FEDOT allows us to use multi-objective optimization only for regression and classification tasks with a limited set of objective functions. After the conducted changes, it can be used for custom tasks and objectives. 

\begin{figure}[ht!]
\centering
\includegraphics[width=.9\columnwidth]{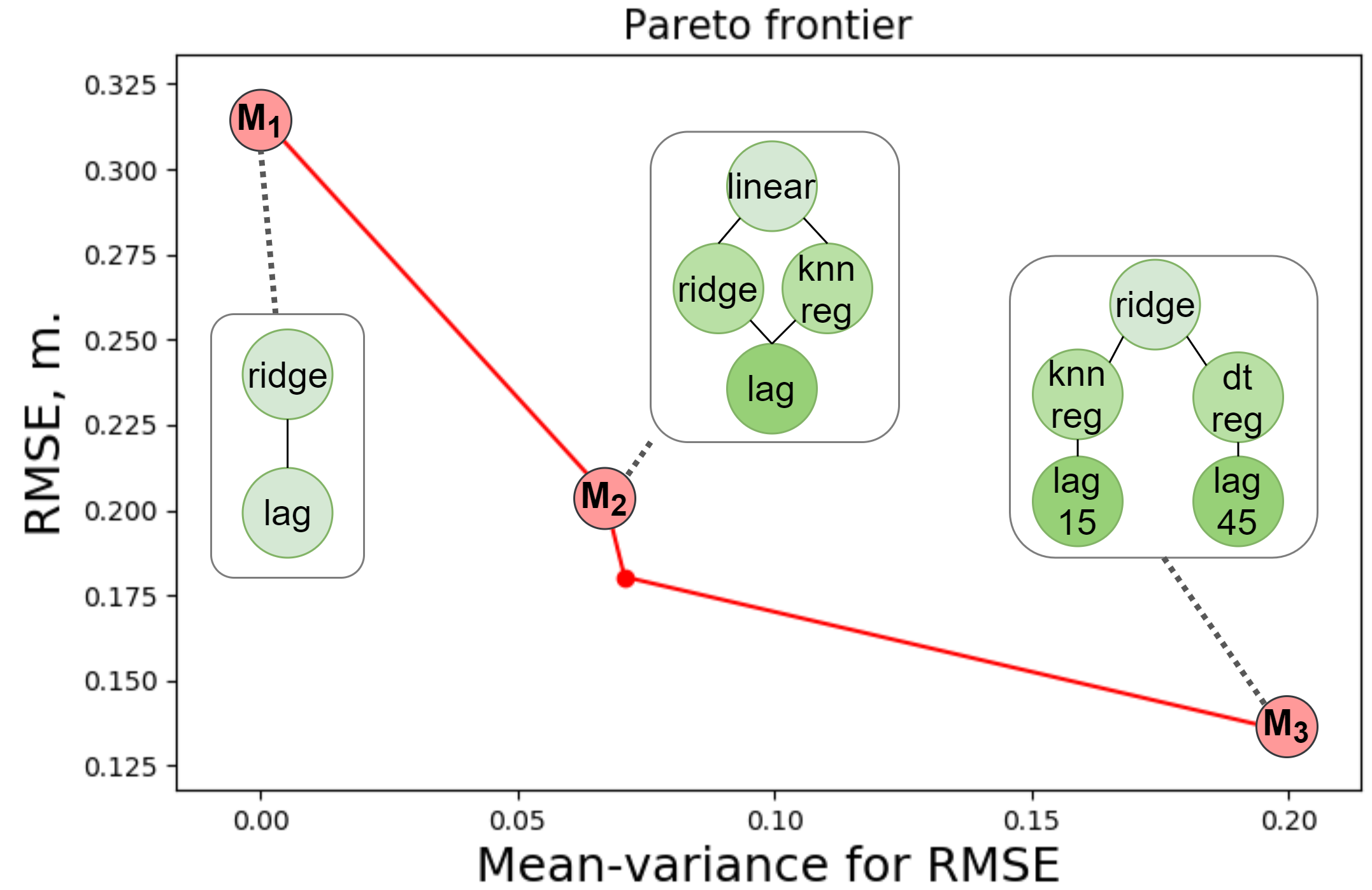}
\caption{The Pareto frontier for the evolutionary multi-objective design of the composite machine learning model. The root mean squared error (RMSE) and mean-variance for RMSE are used as objective functions. The ridge and linear regressions, lagged transformation (referred as lag), k-nearest regression (knnreg) and decision tree regression (dtreg) models are used as a parts of optimal composite model for time series forecasting.}
\label{fig_composite_robust_pareto}
\end{figure}

The generative evolutionary approach is used during the experimental studies to discover the composite machine learning model for the sea surface height dataset. The obtained Pareto frontier is presented is Fig.~\ref{fig_composite_robust_pareto}.

From Fig.~\ref{fig_composite_robust_pareto} we obtain an interpretation that agrees with the practical guidelines. Namely, the structures with single ridge regression ($M_1$) are the most robust, meaning that dataset partition less affects the coefficients. The single decision tree model, on the contrary, the most dependent on the dataset partition model.

\subsection{Closed-form algebraic expressions}
\label{sec:EALG}

The class of the models may include algebraic expressions to obtain better interpretability of the model. As the first example, we present the multi-objective algebraic expression discovery example.

As the algebraic expression, we understand the sum of the atomic functions' products, which we call tokens. Basically token is an algebraic expression with a free parameters (as an example $T=(t;\alpha_1,\alpha_2,\alpha_3)=\alpha_3\sin(\alpha_1 t+\alpha_2)$ with free parameters set $\alpha_1,\alpha_2,\alpha_3$ ), which are subject to optimization. In the present paper, we use pulses, polynomials, and trigonometric functions as the tokens set.

For the mathematical expression models overall, it is helpful to introduce two groups of objectives. The first group of the objectives we refer to as ``quality''. For a given equation $M$, the quality metric $|| \cdot ||$ is the data $D$ reproduction norm that is represented as

\begin{equation}
     Q(M)=|| M-D|| 
\label{eq:norm_quality}
\end{equation}

The second group of objectives we refer to as ``complexity''. For a given equation $M$, the complexity metric is bound to the length of the equation that is denoted as $\#(M)$

\begin{equation}
     C(M)= \#(M)
\label{eq:norm_complexity}
\end{equation}

As an example of objectives, we use rooted mean squared error (RMSE) as the quality metric and the number of tokens present in the resulting model as the complexity metric. First, the model's structure is obtained with a separate evolutionary algorithm to compute the mean squared error. In details it is described in \cite{merezhnikov2020closed}.

\begin{figure}[ht!]
\centering
\includegraphics[width=0.6\linewidth]{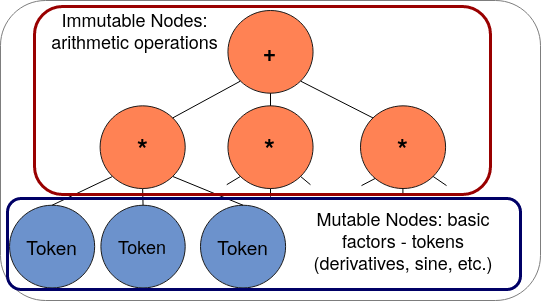}
\caption{The scheme of the composite model, generalizing the discovered differential equation, where red nodes are the nodes, unaffected by mutation or cross-over operators of the evolutionary algorithm. The blue nodes represent the tokens that evolutionary operators can alter.}
\label{fig:ESTAR_structure}
\end{figure}

To perform the single model evolutionary optimization in this case, we make the arithmetic operations immutable. The resulting directed acyclic graph is shown in Fig.~\ref{fig:ESTAR_structure} Thus, the third type of nodes appears - immutable ones. This step is not necessary, and the general approach described above may be used instead. However, it reduces the search space and thus reduces the optimization time without losing generality. 

\begin{figure}[ht!]
\centering
\includegraphics[width=0.8\columnwidth]{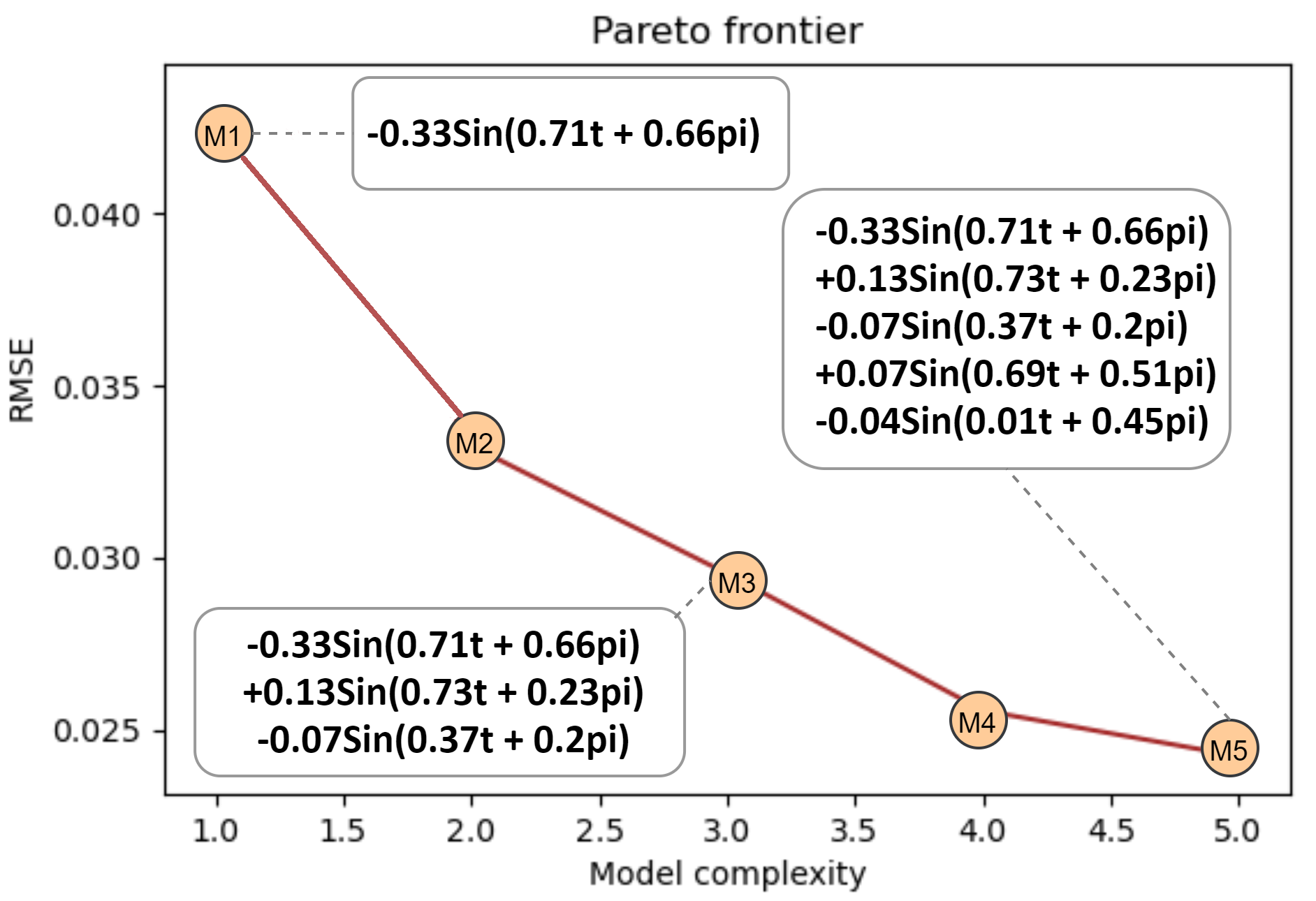}
\caption{The Pareto frontier for the evolutionary multi-objective design of the closed-form algebraic experessions. The root mean squared error (RMSE) and model complexity are used as objective functions.}
\label{fig:ealg_pareto}
\end{figure}

The resulting Pareto frontier for the class of the described class of closed-form algebraic expressions is shown in Fig.~\ref{fig:ealg_pareto}.

Since the origin of time series is the sea surface height in the ocean, it is natural to expect that the closed-form algebraic expression is the spectra-like decomposition, which is seen in Fig.~\ref{fig:ealg_pareto}. It is also seen that as soon as the complexity rises, the additional term only adds information to the model without significant changes to the terms that are present in the less complex model.

\subsection{Differential equations}
\label{sec:EPDE}

The development of a differential equation-based model of a dynamic system can be viewed from the composite model construction point of view. A tree graph represents the equation with input functions, decoded as leaves, and branches representing various mathematical operations between these functions. The specifics of a single equation's development process were discussed in the article \cite{MASLYAEV2021101345}.

The evaluation of equation-based model quality is done in a pattern similar to one of the previously introduced composite models. Each equation represents a trade-off between its complexity, which we estimate by the number of terms in it and the quality of a process representation. Here, we will measure this process's representation quality by comparing the left and right parts of the equation. Thus, the algorithm aims to obtain the Pareto frontier with the quality and complexity taken as the objective functions. 

We cannot use standard error measures such as RMSE since the partial differential equation with the arbitrary operator cannot be solved automatically. Therefore, the results from previous sections could not be compared using the quality metric.

\begin{figure}[ht!]
\centering
\includegraphics[width=0.8\columnwidth]{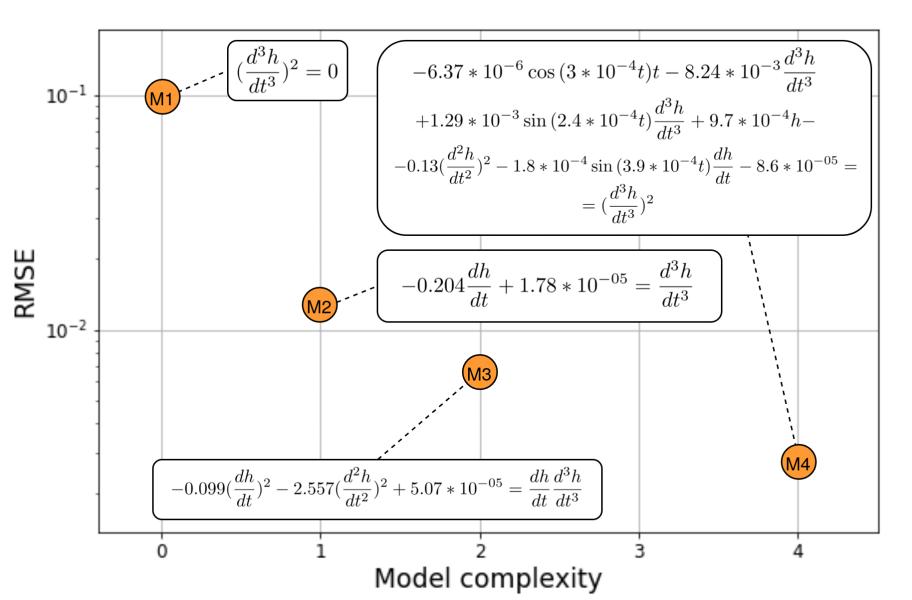}
 \caption{The Pareto frontier for the evolutionary multi-objective discovery of differential equations, where complexity objective function is the number of terms in the left part of the equation, and quality is the approximation error (difference between the left and right parts of the equation).}
\label{fig:eq_pareto}
\end{figure}

Despite the achieved quality of the equations describing the process, presented in Fig.~\ref{fig:eq_pareto}, their predictive properties may be lacking. The most appropriate physics-based equations to describe this class of problems (e.g., shallow-water equations) include spatial partial derivatives that are not available in processing a single time series.    

\section{Conclusion}
\label{sec:conclusion}

The paper describes a multi-objective composite models discovery approach intended for data-driven modeling and initial interpretation.

Genetic programming is a powerful tool for DAG model generation and optimization. However, it requires refinement to be applied to the composite model discovery. We show that the number of changes required is relatively high. Therefore, we are not talking about the genetic programming algorithm. Moreover, the multi-objective formulation may be used to understand how the human-formulated objectives affect the optimization, though this basic interpretation is achieved.

As the main advantages we note:

\begin{itemize}
    \item The model and basic interpretation are obtained simultaneously during the optimization
    \item The approach can be applied to the different classes of the models without significant changes
    \item Obtained models could have better quality since the multi-objective problem statement increases diversity which is vital for evolutionary algorithms
\end{itemize}

As future work, we plan to work on the unification of the approaches, which will allow obtaining the combination of algebraic-form models and machine learning models, taking best from each of the classes: better interpretability of mathematical and flexibility machine learning models.

\section*{Acknowledgements}

This research is financially supported by The Russian Science Foundation, Agreement \#17-71-30029 with cofinancing
of Bank Saint Petersburg.
%
%
%
\bibliographystyle{splncs04}
\bibliography{sample-base}

\end{document}